\documentclass[11pt]{article}

\usepackage{EMNLP2022}
\usepackage{times}
\usepackage{latexsym}
\usepackage{graphicx}
\usepackage{url}[hyphens]
\usepackage{amsmath}
\usepackage{amssymb}
\usepackage{pgfplots}
\usepackage{breqn}
\usepackage{makecell}
\usepackage{tabularx}
\usepackage{booktabs}
\usepackage{multirow}
\usepackage{textcomp}
\usepackage{comment}
\usepackage{bm}
\usepackage{hyperref}
\usepackage{multirow}
\usepackage{microtype}
\usepackage{graphicx}
\usepackage{subcaption}
\usepackage{inconsolata}

\usepackage[utf8]{inputenc}
\usepackage{microtype}
\usepackage{inconsolata}
\title{GersteinLab at MEDIQA-Chat 2023: Clinical Note Summarization from Doctor-Patient Conversations through Fine-tuning and In-context Learning}

\author{
\textbf{Xiangru Tang}$^\spadesuit$ \ \ 
        \textbf{Andrew Tran}$^{\spadesuit}$ \ \ 
                \textbf{Jeffrey Tan}$^{\spadesuit}$ \ \ 
        \textbf{Mark Gerstein}$^\spadesuit$ \ \ 
       \\   
  $^\spadesuit$ Yale
University, New Haven, CT 06520, USA 
  \\   \\
  {\tt \{xiangru.tang, a.tran, jeffrey.tan, mark.gerstein\}@yale.edu}
}
\begin{document}
\maketitle
\begin{abstract}
This paper presents our contribution to the MEDIQA-2023 Dialogue2Note shared task, encompassing both subtask A and subtask B. We approach the task as a dialogue summarization problem and implement two distinct pipelines: (a) a fine-tuning of a pre-trained dialogue summarization model and GPT-3, and (b) few-shot in-context learning (ICL) using a large language model, GPT-4. Both methods achieve excellent results in terms of ROUGE-1 F1, BERTScore F1 (deberta-xlarge-mnli), and BLEURT, with scores of 0.4011, 0.7058, and 0.5421, respectively. Additionally, we predict the associated section headers using RoBERTa and SciBERT based classification models. Our team ranked fourth among all teams, while each team is allowed to submit three runs as part of their submission. We also utilize expert annotations to demonstrate that the notes generated through the ICL GPT-4 are better than all other baselines. The code for our submission is available
~\footnote{https://github.com/gersteinlab/MEDIQA-Chat-2023}.
\end{abstract}

\section{Introduction}
\begin{figure*}[t]
  \centering
  \includegraphics[scale=0.6]{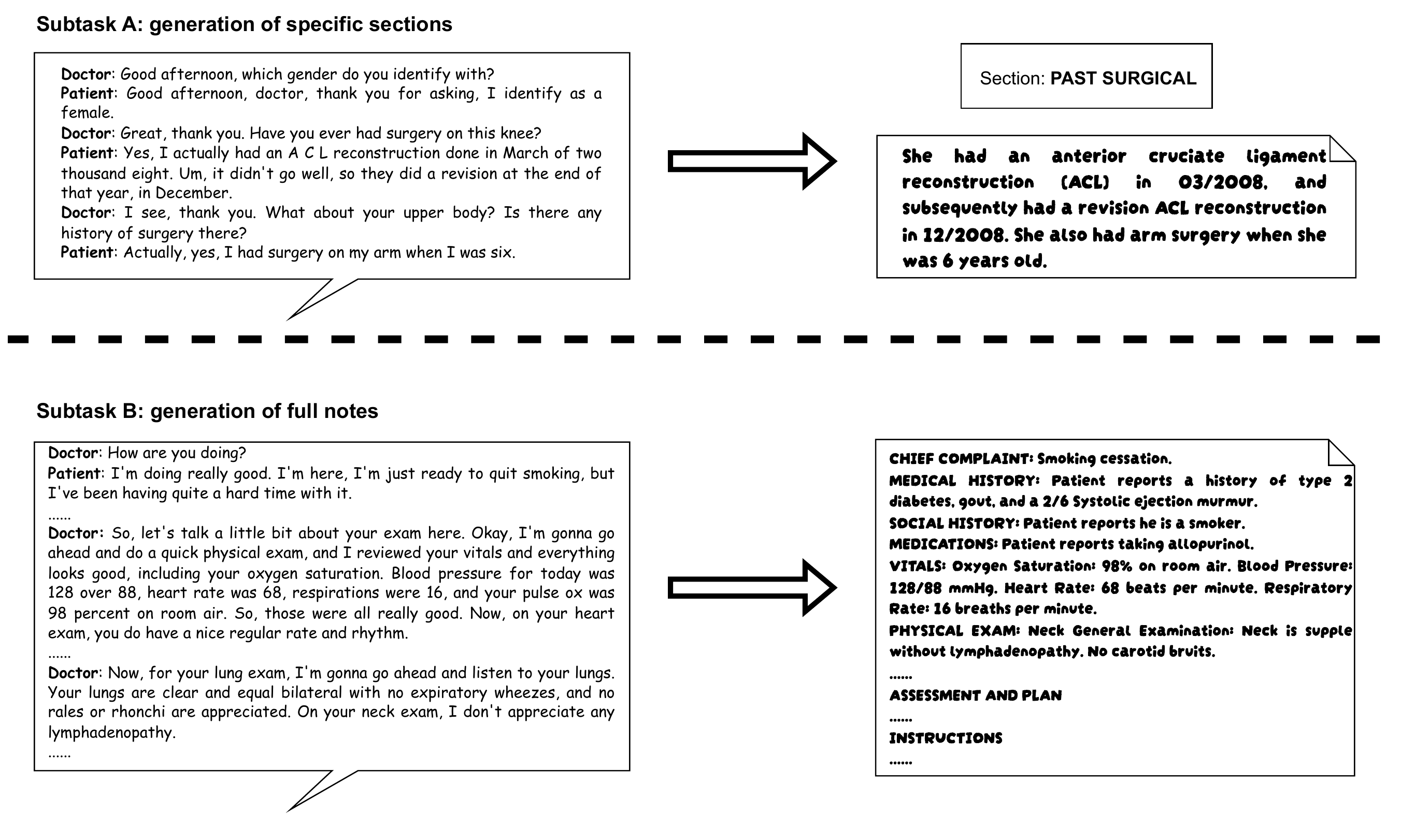}
  \caption{Two examples of our tasks, which include both subtask A and subtask B. In subtask A, the goal is to generate a summary and the corresponding section name for a specific section, while subtask B aims to generate a complete note.}
  \label{fig:example1}
\end{figure*}

The field of medical AI has witnessed significant advancements in recent years, fueled by its promise to transform clinical documentation procedures~\cite{beltagy-etal-2019-scibert,alsentzer2019publicly,huang2019clinicalbert,si2019enhancing,lee2020biobert,gu2021domain}. Extracting clinical notes from doctor-patient interactions is a crucial aspect of maintaining medical records, as it fosters effective communication among healthcare practitioners. By automating this process, healthcare professionals can shift their focus toward patient care and minimize the time dedicated to administrative duties~\cite{jain2022survey,navarro2022few}. The development of efficient and accurate algorithms for summarizing these conversational notes is therefore of paramount importance, as it has the potential to improve overall healthcare quality and efficiency~\cite{quiroz2020identifying,krishna-etal-2021-generating,menon2021deep,michalopoulos2022medicalsum,tang-ali-2023}.

The MEDIQA-Chat 2023 challenge\footnote{https://sites.google.com/view/mediqa2023/clinicalnlp-mediqa-chat-2023} was established to promote the development of novel summarization techniques, specifically targeting the automatic generation of clinical notes from doctor-patient conversations~\cite{mediqa-chat-2023}. The Dialogue2Note and Note2Dialogue shared tasks are designed to stimulate research and innovation in this field, addressing the summarization of medical conversations for clinical note creation and the generation of synthetic doctor-patient conversations for data creation and augmentation. The Dialogue2Note Summarization task entails converting a doctor-patient conversation into a clinical note containing one or multiple note sections, such as Assessment, Past Medical History, or Past Surgical History. This task is subdivided into two subtasks: (A) generating specific sections from conversations~\cite{mts-dialog}, and (B) generating complete notes from conversations~\cite{aci-demo}, two examples are shown in Figure. \ref{fig:example1}.

In this paper, we discuss our submission to both subtask A and subtask B of the shared task:
For \textbf{subtask A}, we first focused on section classification and explored two methods: (1) using RoBERTa~\cite{liu2019roberta} and SciBERT~\cite{beltagy-etal-2019-scibert} with a classification head, and (2) fine-tuning OpenAI's Davinci model~\footnote{https://platform.openai.com/docs/guides/fine-tuning}. Subsequently, we investigated generating specific sections using a fine-tuned pre-trained dialogue summarization model. We employed the CONFIT model~\cite{tang-etal-2022-confit}, which proposes a training strategy that enhances the factual consistency and overall quality of summaries through a novel contrastive fine-tuning approach.
For \textbf{subtask B}, we examined how to utilize large language models (LLMs) like GPT. We (1) fine-tuned OpenAI's Davinci model and (2) explored in-context learning~\cite{dong2022survey} with GPT-4~\footnote{https://openai.com/research/gpt-4}. We achieved promising results on automated metrics (ROUGE, BERTScore~\cite{zhang2019bertscore}, and BLEURT~\cite{sellam-etal-2020-bleurt}), and our outcomes were also assessed manually. Although the GPT-based model scored slightly lower on automated metrics, it received high scores in human evaluations. We believe that for zero-shot models, existing automated metrics may not be the most appropriate evaluation method, suggesting a potential direction for future research.

\section{Tasks}
\subsection{Task Formulation}
In this paper, we focus solely on the Dialogue2Note Summarization task of MEDIQA-Chat Tasks @ ACL-ClinicalNLP 2023. The main tasks include:

\begin{itemize}
\item \textbf{Dialogue2Note Summarization}: Given a doctor-patient conversation, participants are required to generate a clinical note summarizing the conversation, including one or multiple note sections (e.g., Assessment, Past Medical History, Past Surgical History). This task comprises two subtasks:
\begin{itemize}
\item \textbf{Subtask A}: Generating specific sections from doctor-patient conversations~\cite{mts-dialog}.
\item \textbf{Subtask B}: Generating full notes from doctor-patient conversations~\cite{aci-demo}.
\end{itemize}

\item \textbf{Note2Dialogue Generation}: Participants are tasked with generating a synthetic doctor-patient conversation based on the information described in a given clinical note~\cite{aci-demo}.
\end{itemize}

For \textbf{subtask A}, the training set consists of 1,201 pairs of conversations and associated section headers and contents, while the validation set includes 100 pairs of conversations and their summaries. A full list of normalized section headers is provided in the paper.

As for \textbf{subtask B}, the training set is composed of 67 pairs of conversations and full notes, and the validation set includes 20 pairs of conversations and clinical notes.

Lastly, the \textbf{Note2Dialogue Generation task}'s training set comprises 67 pairs of full doctor-patient conversations and notes, with the validation set containing 20 pairs of full conversations and clinical notes. The Task-A training and validation sets (1,301 pairs) could be used as additional training data.

Thus, we could formally define the tasks as follows. Given a doctor-patient conversation $C = \{c_1, c_2, ..., c_n\}$, 
where $c_i$ represents the $i^{th}$ utterance in the conversation 
and $n$ denotes the total number of utterances, the goal of the 
Dialogue2Note Summarization task is to generate a clinical note 
summarizing the conversation.

\begin{figure}[htbp!]
  \centering
  \begin{subfigure}{0.3\textwidth}
    \includegraphics[width=\textwidth]{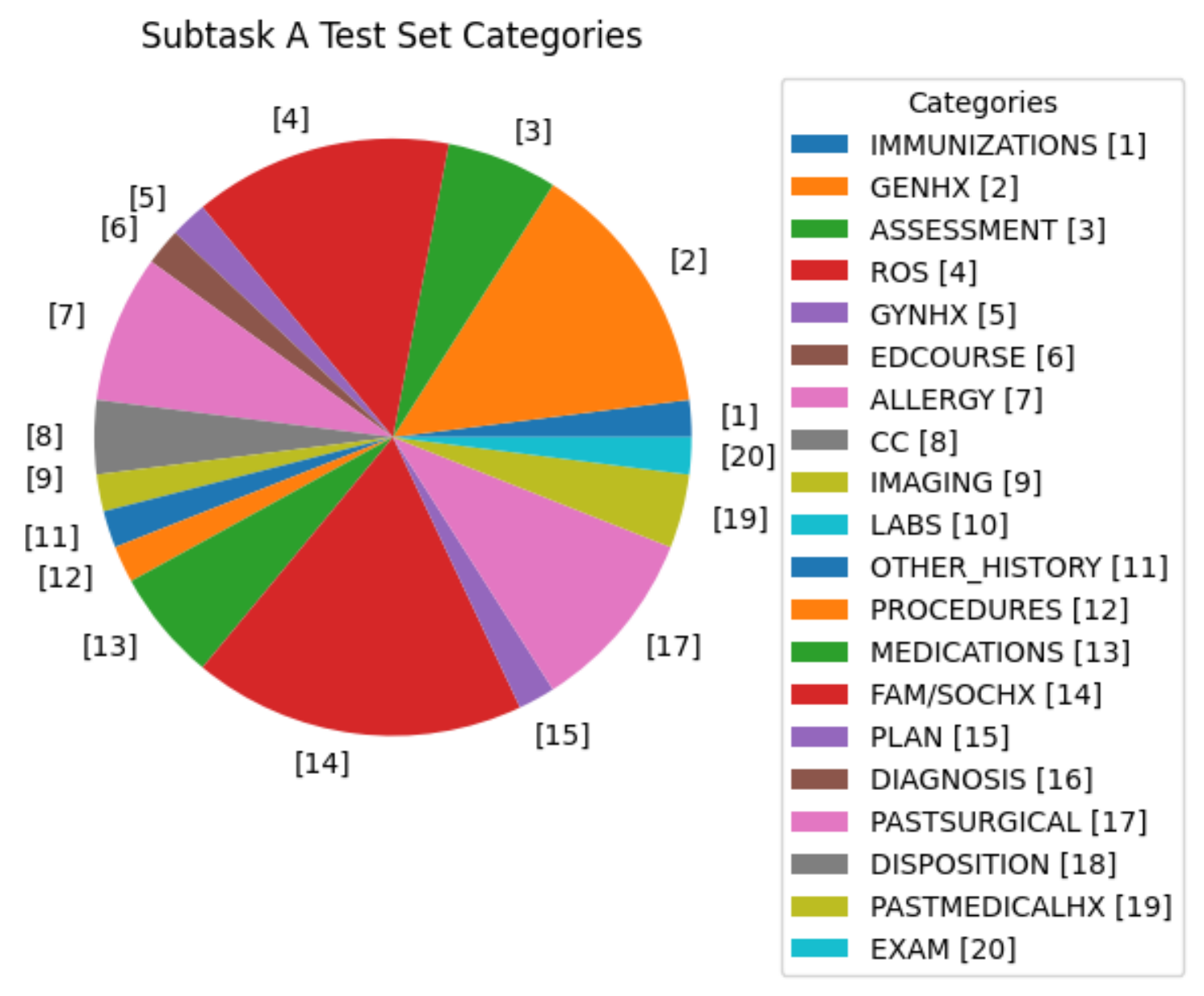}
  \end{subfigure}
  \hfill
  \begin{subfigure}{0.3\textwidth}
    \includegraphics[width=\textwidth]{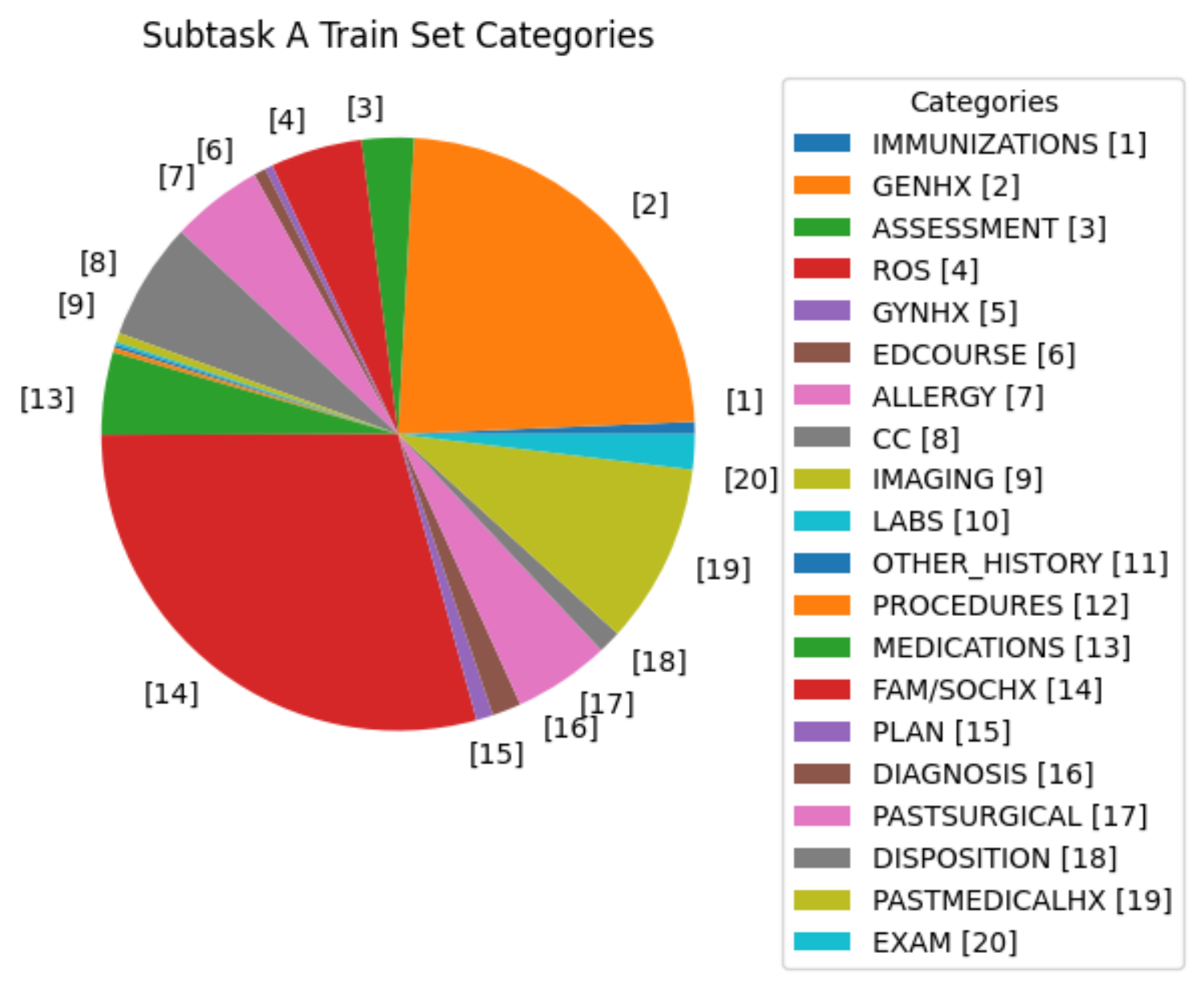}
  \end{subfigure}
  \hfill
  \begin{subfigure}{0.3\textwidth}
    \includegraphics[width=\textwidth]{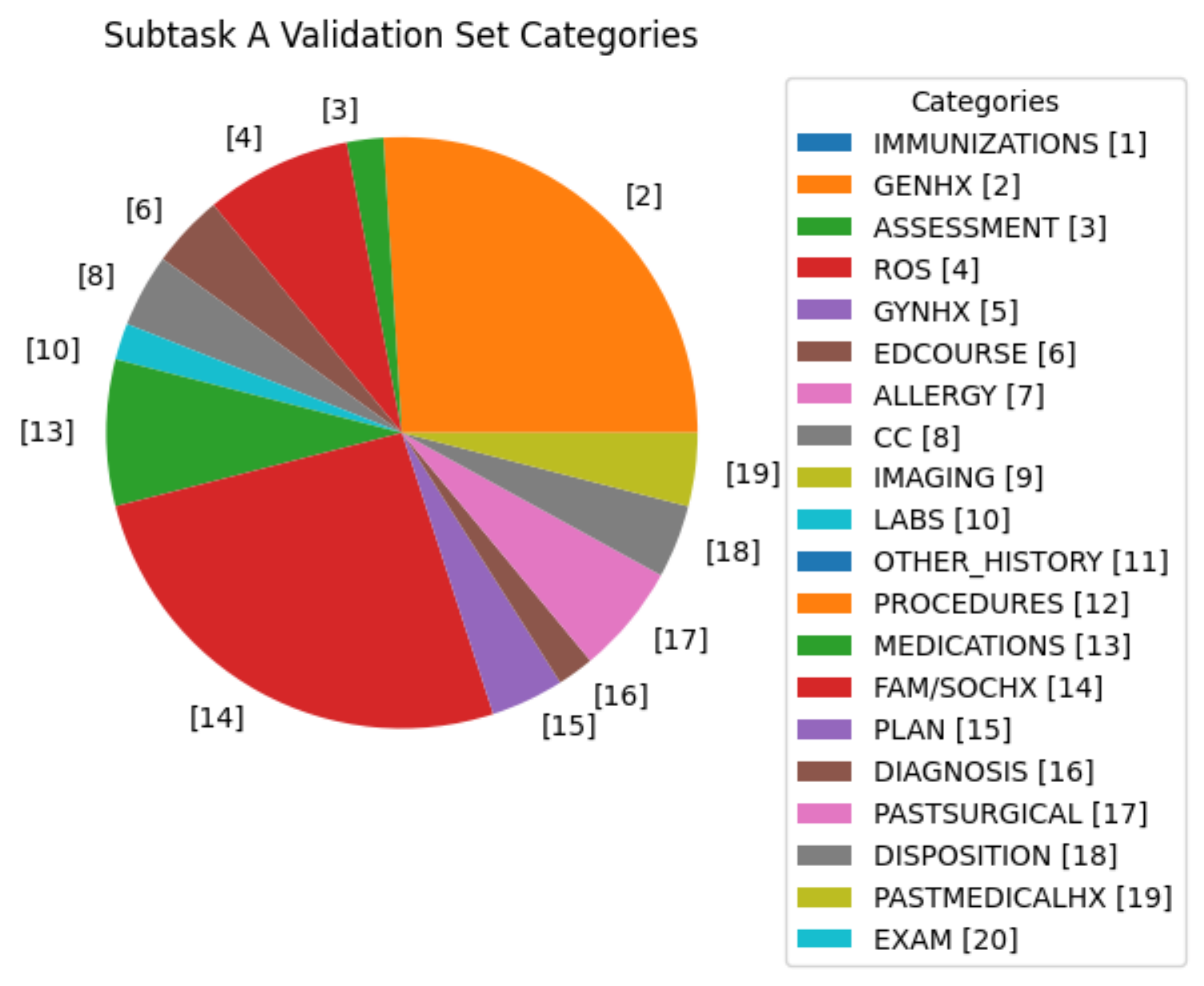}
  \end{subfigure}
  \caption{The proportion of section categories in subtask A.}
  \label{fig:three_pdfs}
\end{figure}

For \textbf{subtask A}, the objective is to generate a specific section 
summary $S_j$ and its corresponding section header $H_j$ for a 
given conversation $C$. The output can be represented as a 
tuple $(H_j, S_j)$. For \textbf{subtask B}, the goal is to generate a complete clinical note 
$N = \{S_1, S_2, ..., S_m\}$, where $S_i$ represents the $i^{th}$ 
summary section and $m$ denotes the total number of sections. 
Each section $S_i$ is associated with a section header $H_i$.
We use a combination of various evaluation 
metrics such as ROUGE, BERTScore, and BLEURT.

\subsection{Data Analysis}

\begin{figure}[htbp!]
  \centering
  \includegraphics[scale=0.5]{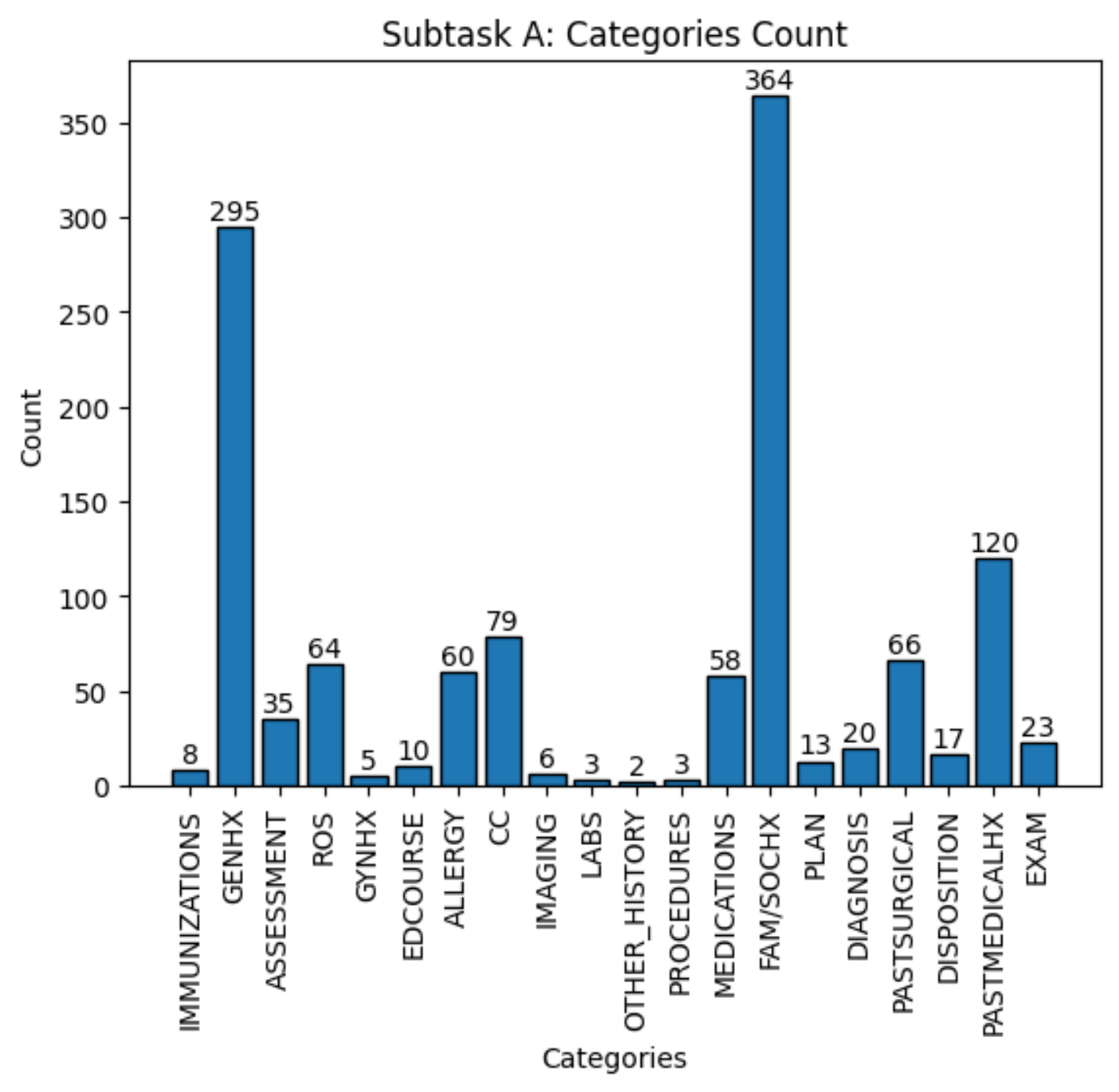}
  \caption{The numbers of section categories in subtask A. Here are the totals on the train and validation sets.}
  \label{fig:example:section}
\end{figure}

For the section classification task in Subtask A, we created pie charts (See Figure. \ref{fig:three_pdfs}) representing the proportions of different sections in the train, test, and validation sets to analyze the distribution differences among them. We observed that there is no significant gap in the section categories across the data splits. However, there is a considerable disparity in the number of instances among different categories~\ref{fig:example:section}, with some sections having very few data points. This may lead to insufficient training and poor performance for those underrepresented categories.

\begin{figure}[hb!]
  \centering

    \begin{subfigure}[b]{0.2\textwidth}
    \includegraphics[width=\textwidth]{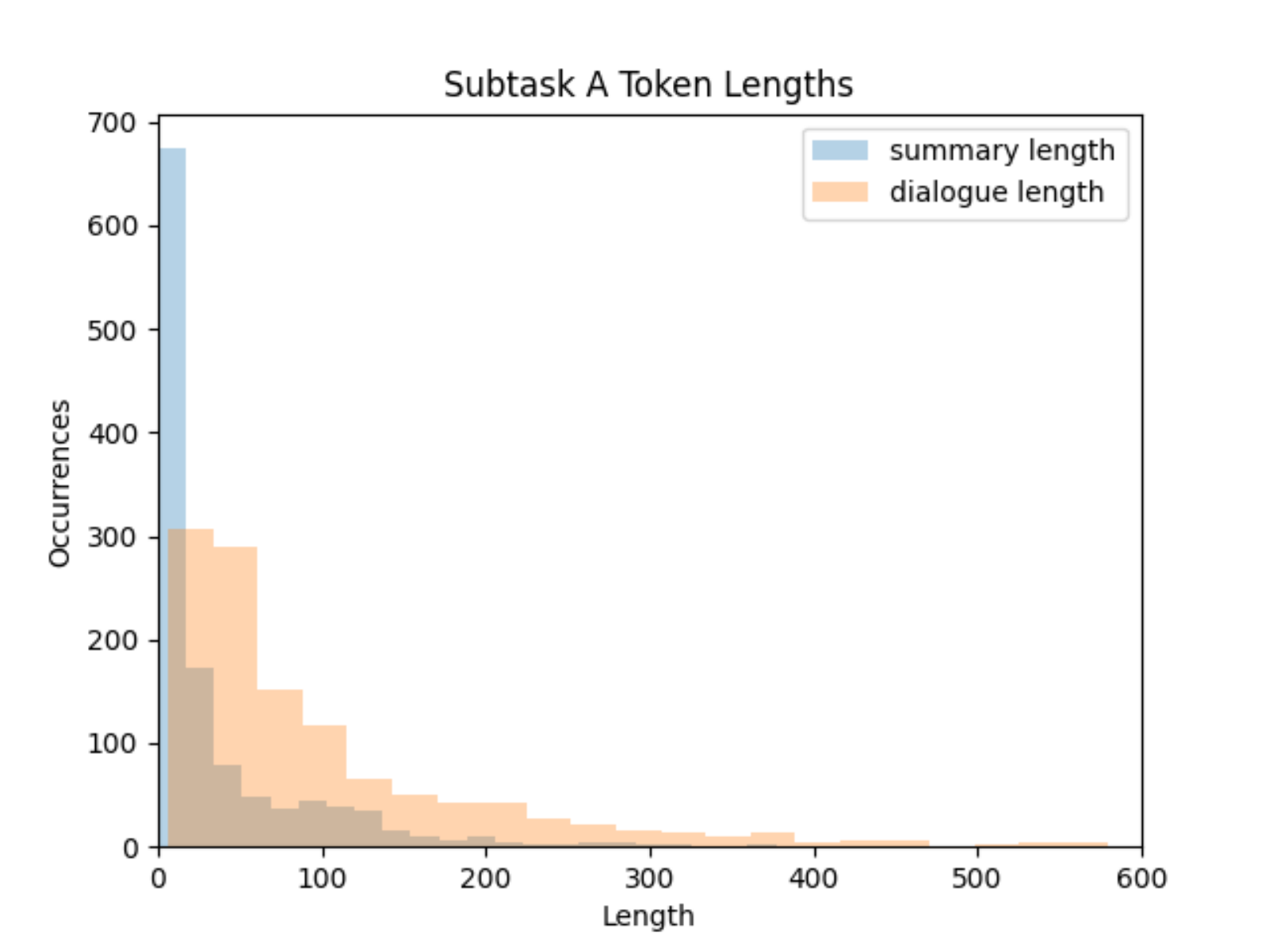}
  \end{subfigure}
  \begin{subfigure}[b]{0.2\textwidth}
    \includegraphics[width=\textwidth]{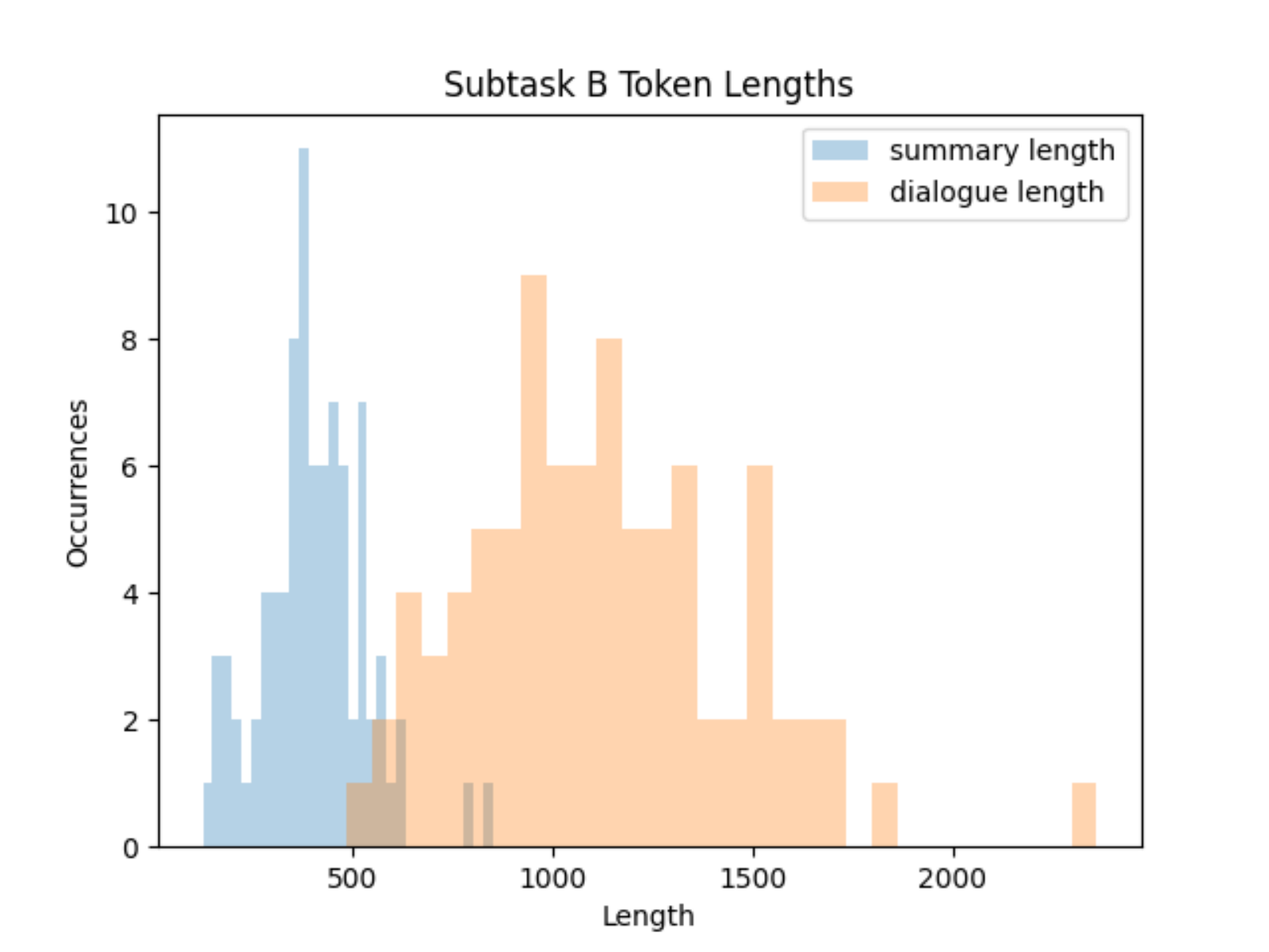}
  \end{subfigure}
  
  \caption{Histogram of token lengths for subtask A train and validation sets in subtask A and B.}
  \label{fig:example:tokenlengthsA}
\end{figure}

\begin{figure}[htbp!]
    \centering
    \begin{subfigure}[b]{0.2\textwidth}
        \centering
        \includegraphics[width=\textwidth]{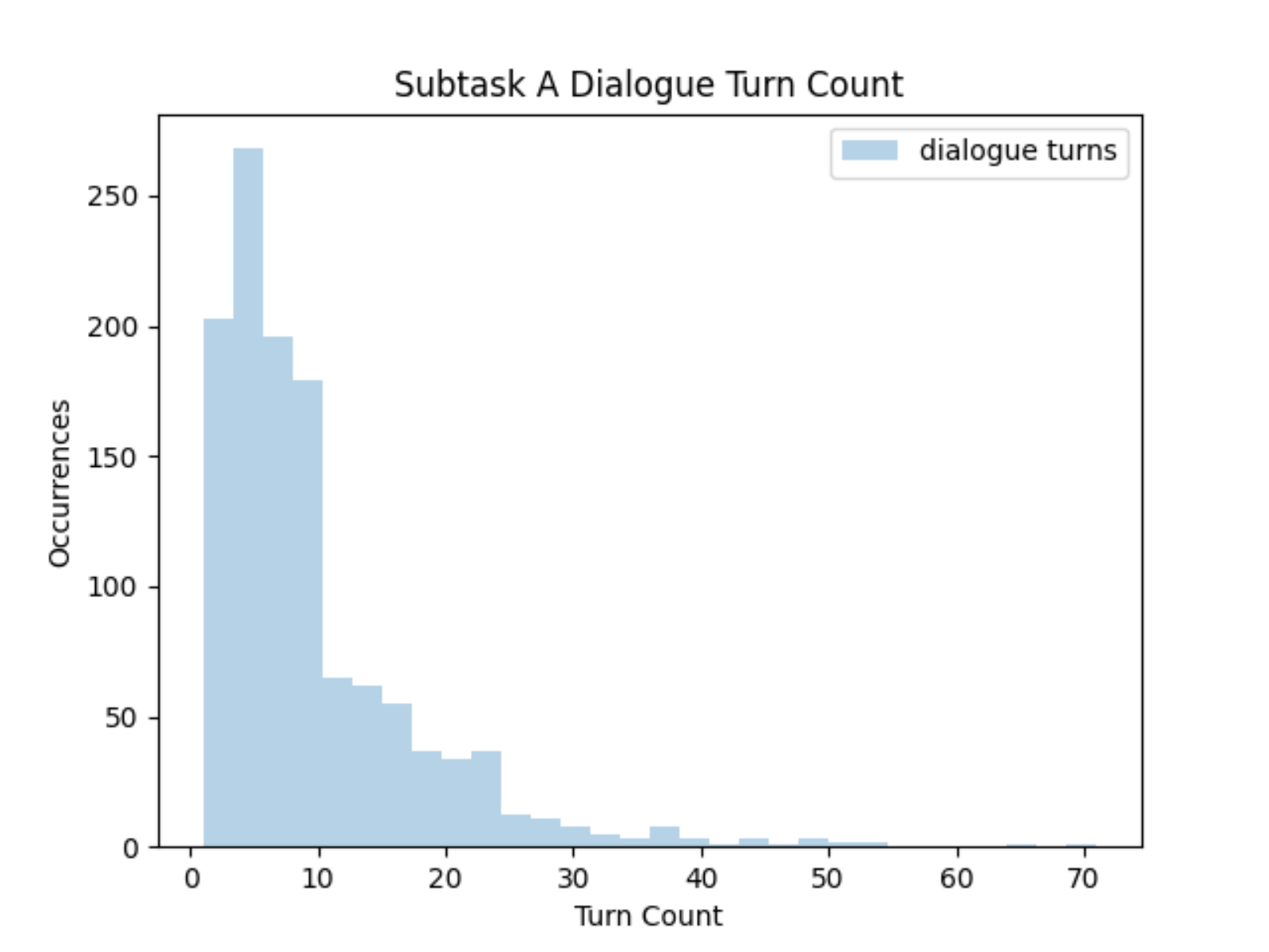}
    \end{subfigure}%
    \begin{subfigure}[b]{0.2\textwidth}
        \centering
        \includegraphics[width=\textwidth]{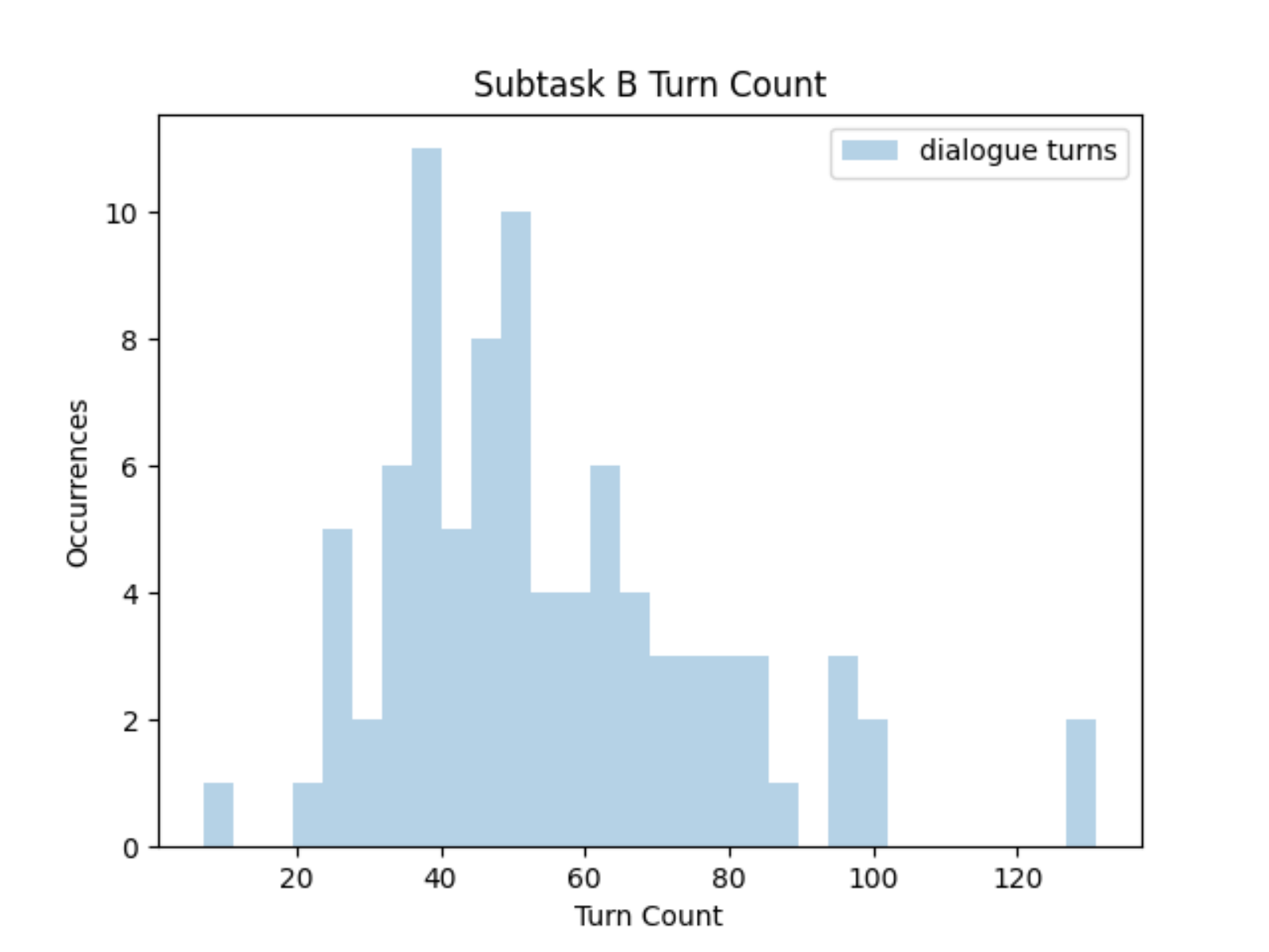}
    \end{subfigure}
    \caption{Histogram of utterance numbers for subtask A train and validation sets in subtask A and B.}
    \label{fig:diaTurns}
\end{figure}

\begin{figure}[htbp!]
  \centering

    \begin{subfigure}[b]{0.2\textwidth}
    \includegraphics[width=\textwidth]{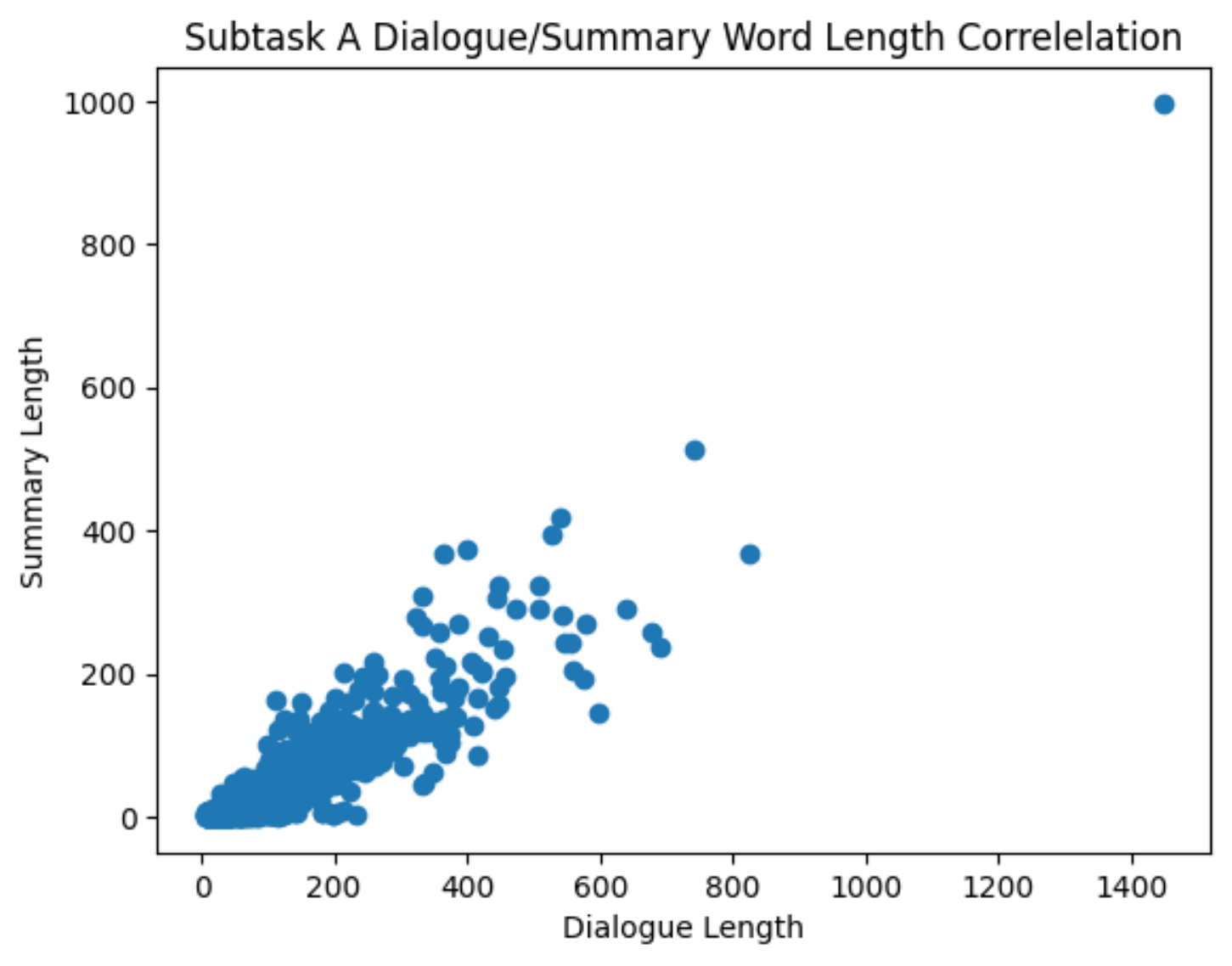}
  \end{subfigure}
  \begin{subfigure}[b]{0.2\textwidth}
    \includegraphics[width=\textwidth]{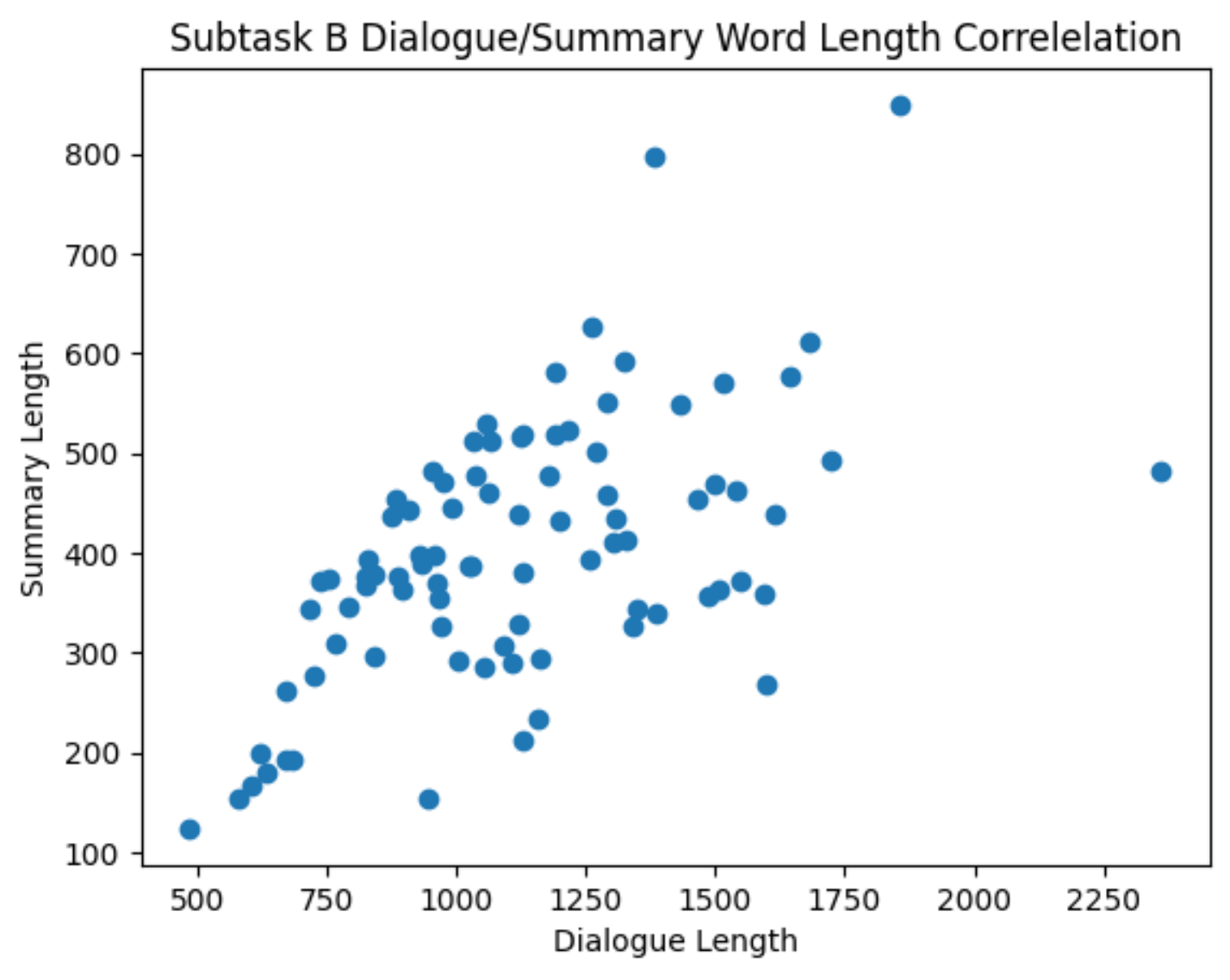}
  \end{subfigure}
  
  \caption{Dialogue Length plotted against Summary Length for each data entry. The graphs show a noticeable positive correlation.}
  \label{DiaSumScatterB}
\end{figure}

In Figure. \ref{fig:example:tokenlengthsA}, we illustrate the length of input dialogues and the length of output summaries. And Figure.   \ref{fig:diaTurns} shows the number of utterances.
In the picture. \ref{DiaSumScatterB}, the length of the dialogue is plotted against the length of the summary for each data entry. The graphs show a noticeable positive correlation, thus indicating that longer dialogues do have significantly longer summaries. Additionally, we can see that dialogue lengths increase at about twice the rate of summary lengths, so our summaries should be about one-third to one-half the dialogue length.

\section{Methodology}

\begin{figure}[t]
  \centering
  \includegraphics[scale=0.6]{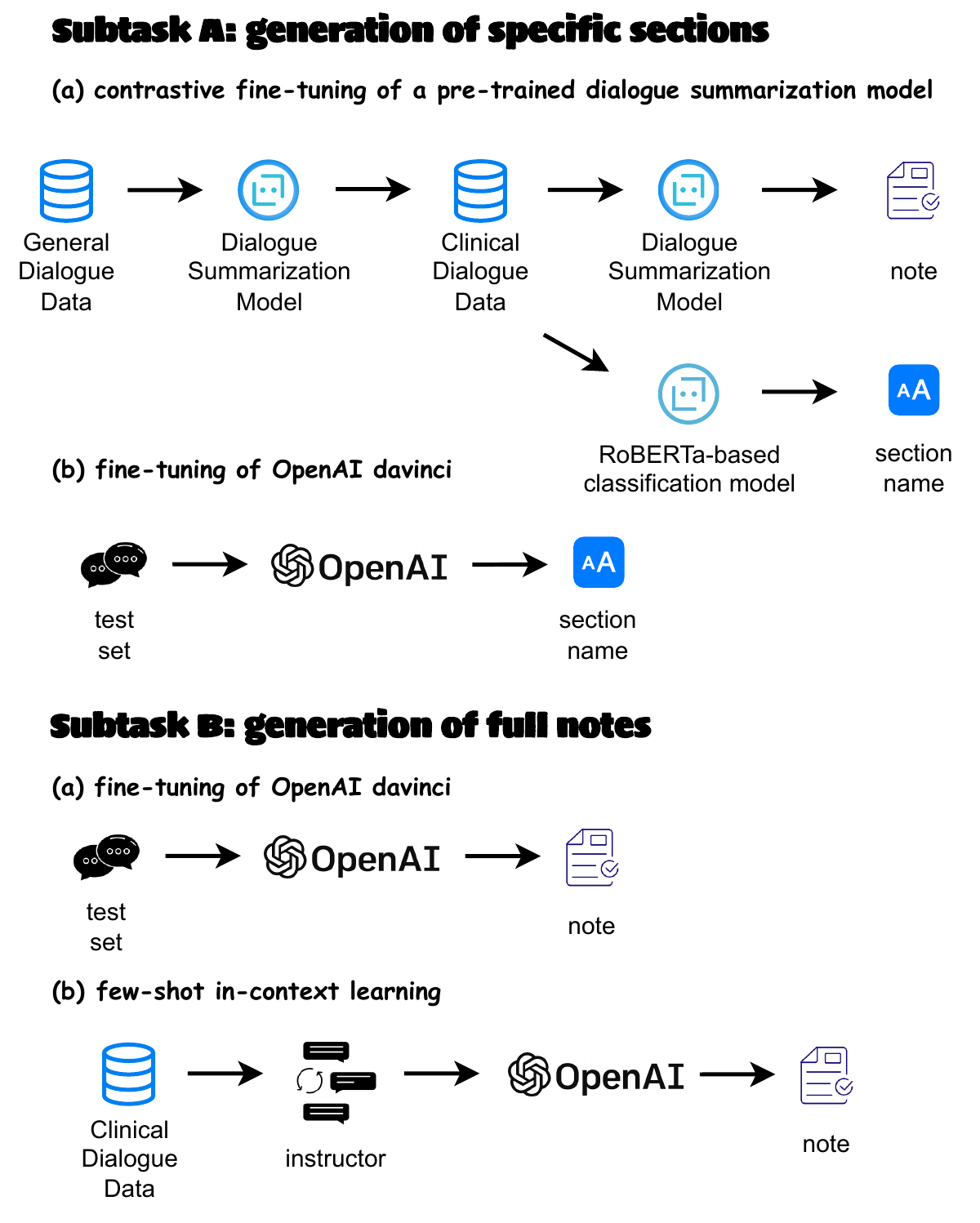}
  \caption{This diagram illustrates the pipeline we employed for both subtask A and subtask B. For subtask A, we fine-tune a dialogue summarization model to generate notes, and to generate section names, we use a RoBERTa model and fine-tuning of OpenAI's Davinci model. For subtask B, we perform fine-tuning as well as few-shot in-context learning to achieve our desired results.}
  \label{fig:example}
\end{figure}
In this section, we will describe how we employed various approaches to perform subtask A and subtask B. We will discuss each subtask in detail, outlining the methods used and the rationale behind our choices to achieve optimal results.

\begin{table*}[t]
\centering
\small
\begin{tabular}{c|c|c|c|c|c|c|c|c}
    \toprule
    \textbf{Models}          & \textbf{R\textsubscript{1}} & \textbf{R\textsubscript{2}} & \textbf{R\textsubscript{L}} & \textbf{R\textsubscript{Lsum}} & \textbf{BERTScore\textsubscript{p}} & \textbf{BERTScore\textsubscript{r}} & \textbf{BERTScore\textsubscript{f1}} & \textbf{BLEURT}  \\ \midrule
    CONFIT   & 0.3882 & 0.1966 & 0.3214 & 0.3214    & 0.7037             & 0.7065            & 0.7           & 0.5294           \\ 
        CONFIT\textsubscript{dynamic}   & 0.4011 & 0.2147 & 0.3322 & 0.3322    & 0.7115             & 0.7102            & 0.7058        & 0.5421            \\ 
    \bottomrule
\end{tabular}
\caption{Comparison of performance between the CONFIT model and the CONFIT model with dynamic max length on various evaluation metrics.}
\label{table:performance}
\end{table*}

\subsection{Subtask A}

We define this task as a dialogue summarization problem and, therefore, we selected a state-of-the-art dialogue summarization model, CONFIT~\cite{tang-etal-2022-confit}, as our foundation to fine-tune. Firstly, CONFIT is based on BART and has been fine-tuned on the SAMSum dialogue summarization dataset. We utilized the model fine-tuned on SAMSum and further fine-tuned it on MEDIQA subtask A data to generate notes. It is worth noting that ideally, we should have added an additional step, fine-tuning the model on PubMed summarization data before fine-tuning on subtask A data, which would enable the model to better understand the clinical summarization task. We set max-input-length to 1024 and keep output max-length at 128. For generating section names, we used a RoBERTa model and fine-tuning of OpenAI's Davinci model. We fine-tuned RoBERTa to classify text based on 20 predefined categories. Additionally, we invoked OpenAI's API and used a customized model (davinci:ft-personal-2023-03-15-08-22-14) to classify the dialogues. This model classifies text according to the 20 predefined categories, with the classification implemented through the OpenAI API.

\subsection{Subtask B}

We explored two approaches for utilizing OpenAI's large-scale language models. First, we defined a function to shorten the dialogue, ensuring it does not exceed the maximum token length of 1200. Next, we used another function to call OpenAI's API and employ a customized model (davinci:ft-personal-2023-03-23-05-58-11) to generate the corresponding notes. We fine-tuned the Davinci model and, during this process, implemented multiple generation attempts to adjust the maximum token length, setting max tokens to 800. We experimented with temperatures of 0.0 and 0.2. We used the following prompt: "Please summarize the following dialog between doctors and patients from the perspective of the doctor, and be sure to include all important details about the patient."

Moreover, we employed in-context learning, choosing GPT-4~\footnote{https://platform.openai.com/docs/models/gpt-4} and designing a prompt that included natural language instructions and context examples. We limited the input prompt length to 6000 tokens and the output length to 2000 tokens. We used two contexts per instance and set the temperature parameter to 0.2. In our prompt template, we incorporate three main components: instructions, in-context examples, and test input dialogue. The final "FULL NOTE:" indicates the output the model needs to generate.
Our instructions are as follows: "Write a clinical note for this doctor-patient dialogue. Use the example notes below to know the different sections." This guides the model to generate a clinical note based on the given doctor-patient dialogue, taking into account the structure and sections observed in the provided examples.

\section{Results}

We report the accuracy of the section classification. For note generation in subtask A, we present the following metrics: Rouge-1, Rouge-2, Rouge-L, Rouge-Lsum, BERTScore precision~\cite{zhang2019bertscore}, BERTScore recall~\cite{zhang2019bertscore}, BERTScore F1~\cite{zhang2019bertscore}, BLEURT~\cite{sellam-etal-2020-bleurt}, and aggregate score. The aggregate score is the arithmetic mean of ROUGE-1 F1, BERTScore F1, and BLEURT-20 (Pu et al., 2021). 

Additionally, we implemented a dynamic max length feature.For the CONFIT model and the CONFIT model with dynamic max length, we obtained aggregate scores of 0.5392 and 0.5497, respectively. The results demonstrate that the dynamic max length approach slightly improves the overall performance of the model in the summarization task. Initially, it encodes the input dialogue into a vector using a pre-trained tokenizer. Subsequently, it dynamically calculates the maximum length of the summary to be generated, based on the length of the dialogue. The formula employed here is $max_{length} = round(0.55 * dialog_{len} + 18)$, which determines the maximum length of the summary according to the dialogue's word count. Then, the BART model generates the summary while controlling the max length of the output. Ultimately, the generated summary is decoded into text.

The results in Table \ref{table:performance} show a comparison between the performance of the CONFIT model and the CONFIT model with dynamic max length on various evaluation metrics, including ROUGE scores (R\textsubscript{1}, R\textsubscript{2}, R\textsubscript{L}, and R\textsubscript{Lsum}), BERTScore (BERTScore\textsubscript{p}, BERTScore\textsubscript{r}, and BERTScore\textsubscript{f1}), and BLEURT. The CONFIT\textsubscript{dynamic} model achieves better performance on most evaluation metrics, suggesting that the incorporation of dynamic max length improves the overall quality of the generated summaries.
\begin{table}[!]
\centering
\begin{tabular}{@{}lc@{}}
\toprule
\textbf{Models} & \textbf{Accuracy} \\ \midrule
OpenAI davinci & 0.745 \\
SciBERT & 0.710 \\
RoBERTa & 0.700 \\ \bottomrule
\end{tabular}
\caption{Accuracy of section classification for different models.}
\label{table:section}
\end{table}

In Table \ref{table:section}, we present the accuracy of section classification for three different models, OpenAI davinci, SciBERT, and RoBERTa. The task involves classifying the sections into one of the 20 pre-defined categories. As shown in the table, OpenAI davinci achieves the highest accuracy of 0.745, outperforming both SciBERT and RoBERTa, which achieve accuracies of 0.710 and 0.700, respectively. This indicates that the OpenAI davinci model is more effective in classifying sections in this 20-class classification task. In addition to the RoBERTa classifier, we also utilized SciBERT to improve the initial classification performance. By incorporating the domain-specific knowledge embedded in SciBERT, we were able to enhance the accuracy of our section classification task.

\begin{table}[!]
\centering
\begin{tabular}{@{}lcccc@{}}
\toprule
\textbf{Models} & \textbf{R\textsubscript{1}} & \textbf{R\textsubscript{2}}& \textbf{R\textsubscript{L}} & \textbf{R\textsubscript{Lsum}} \\ \midrule
ICL & 0.5821& 	0.3209& 	0.4032	& 0.5443
 \\
Davinci (t=0.0) & 0.5008	&0.2506	&0.3282	&0.4668 \\
Davinci (t=0.2) & 0.5004 &	0.2502&	0.3249&	0.4675 \\ \bottomrule
\end{tabular}
\caption{The evaluation of the ROUGE scores for subtask 2 full note summarization.}
\label{table:subtaskb}
\end{table}

Table \ref{table:subtaskb} presents the evaluation of the ROUGE scores for subtask 2 full note summarization. The table includes the ROUGE-1, ROUGE-2, ROUGE-L, and ROUGE-Lsum scores for each model. Higher ROUGE scores indicate better summarization quality, reflecting the extent to which the generated summaries capture the important information and coherence of the full notes.
Except for our submission, we expanded our approach by incorporating in-context learning. As the capabilities of large language models (LLMs) continue to advance, in-context learning (ICL) has emerged as a novel paradigm in the field of natural language processing (NLP). In ICL, LLMs make predictions based on a limited set of examples, augmented with additional context. This approach allows LLMs to leverage the power of context in order to enhance their predictive abilities. By incorporating contextual information during the learning process, LLMs are able to generate more accurate and contextually relevant predictions. In the evaluation, the in-context model achieves the highest ROUGE scores, indicating that it generates summaries that have a higher overlap and alignment with the reference summaries of the full notes. Comparatively, the "Davinci" model with a temperature of 0.0 achieves lower ROUGE scores, but still performs better than the "Davinci" model with a temperature of 0.2 in terms of ROUGE scores.

Based on previous studies that have shown that automated metrics are not suitable for evaluating the results generated by zero-shot models~\cite{goyal2022news}, we sought alternative evaluation methods. To address this, we enlisted the expertise of three medical students to manually rate 50 note summarization outputs. Following established practices~\cite{tang-etal-2022-investigating}, we employed a scoring system of 1-10 to assess the quality of the generated summaries. This expert evaluation provides a more comprehensive and reliable assessment of the performance of our models in Subtask A and Subtask B, capturing nuanced aspects that automated metrics may not capture accurately. By incorporating expert assessments, we aim to enhance the evaluation process and gain deeper insights into the capabilities and limitations of our zero-shot models in the context of note summarization.
In this study, we also implemented traditional fine-tuned summarization models such as BART~\cite{lewis2019bart}, T5~\cite{raffel2020exploring}, and PEGASUS~\cite{zhang2020pegasus} as baselines. However, it is important to note that due to the absence of a test set reference, we were unable to compute automated metrics for evaluation. Therefore, we solely rely on the results of the manual assessments conducted by human evaluators. Although automated metrics are commonly used to evaluate summarization models, the absence of a reference necessitates a shift towards expert judgments to assess the quality and effectiveness of the generated summaries. Our focus is on reporting the outcomes of the manual evaluations as a reliable measure of the performance of the models.

\begin{table}[!]
\centering
\begin{tabular}{@{}lc@{}}
\toprule
\textbf{Models} & \textbf{Score} \\ \midrule
Subtask A BART  & 7.2 \\
Subtask A T5  &6.6 \\
Subtask A PEGASUS  & 7.0 \\
Subtask A CONFIT  & 7.2 \\
Subtask A  CONFIT\textsubscript{dynamic}   &7.8 \\\midrule
Subtask B BART & 3.5 \\
Subtask B T5 & 3.8 \\
Subtask B PEGASUS & 3.2 \\
Subtask B ICL & 6.6 \\
Subtask B Davinci (t=0.0) & 5.3 \\
Subtask B Davinci (t=0.2) & 5.2 \\
\bottomrule
\end{tabular}
\caption{Expert annotation of generated notes in a scale of 1-10 on subtask A and B.}
\label{table:human}
\end{table}

The results demonstrate that, in subtask A, CONFIT outperforms other baseline methods. In subtask B, it is evident that traditional models do not perform well, and we believe the main difference lies in the input length. As all the examples exceed the maximum input length limitation of traditional models, and the reference also significantly surpasses the maximum length limitation for generation, it is naturally challenging to generate ideal notes. Therefore, in terms of human evaluation, both fine-tuning OpenAI models and utilizing ICL perform far better than using traditional fine-tuned models. The superior performance is likely attributed to OpenAI's models, which provide longer input and output limitations.

\section{Limitations and Discussion}
We would like to discuss some of our findings and thoughts here. The MEDIQA-Chat 2023 challenge indeed presents an excellent opportunity for us to reflect on existing models, analyze their strengths and weaknesses, and investigate their performance on the task of clinical note generation from doctor-Patient conversations.

\paragraph{Evaluation} Automatic metrics often do not have a strong correlation with the quality of summaries, especially as there are no readily available automatic evaluation metrics specifically designed for zero-shot or few-shot LLM evaluation. Almost all models on Subtask A have similar results of automatic evaluation. Therefore, human evaluation becomes an essential component for accurately assessing the quality of notes generated by LLMs. There is a need to develop more effective automatic evaluation methods for a zero-shot/few-shot generation or super-long output.

\paragraph{Length Limitation of Subtask A} One significant limitation of our Transformer-based model is that it does not directly consider length during its generation process. This often results in the production of overly verbose summaries. In contrast, models like OpenAI's GPT-3 and GPT-4 have much longer input and output limitations, allowing them to handle more extensive text samples more effectively. It is worth noting that our model was trained on the SAMSum dataset, which has longer texts compared to subtask A. Consequently, our model struggles to adapt to the shorter length requirements of subtask A. Moreover, the training dataset for subtask A is relatively small, which further complicates the model's adaptation. Future exploration should look at how to constrain the conciseness of generated summaries, which may involve reconsidering the generation method chosen or examining other techniques to promote brevity. Developing methods to better control the length of generated summaries is essential to improve their relevance, coherence, and usability in real-world applications.

\paragraph{Length Limitation of Subtask B} For Subtask B, it is challenging to achieve reasonable results using fine-tuned models. In reality, this task is more representative of real-world scenarios, where inputs and outputs are considerably long, and the output is expected to maintain a specific structure and format. Thus, we see the main advantage of using contextual examples lies in their ability to guide the structure, style, and length of the desired output. We believe that OpenAI's LLM is well-suited for similar real-life scenarios, provided that it is given an appropriate context. In such cases, its performance will significantly surpass that of fine-tuned Transformer-based models.

\paragraph{Factual inconsistentcy} While our study did not specifically investigate the following issues, we noted several factual errors that occur in summaries.
A previous study has shown that LLMs also exhibit a noticeable occurrence of attribute errors and misinterpretation errors~\cite{tang2023evaluating}.

\paragraph{Prompting } We find that prompt template and demonstration example selection both have a substantial impact on results. Using more prompt examples for demonstration improves significantly. We acknowledge that we did not explore different selection strategies, such as SemScore, LMScore, and TLength, which involve using top-ranked examples. These strategies have been shown to potentially improve the quality of the generated summaries by selecting more effective prompt examples. While our current approach did not incorporate these strategies, we recognize that exploring and incorporating better prompt examples could potentially yield improved results. This is an area that warrants further investigation and experimentation to enhance the performance of our models in future iterations of the study.

\paragraph{Data Privacy} Both GPT-3 and GPT-4 are not local models; we utilize OpenAI's API to run these models, which actually violates data protection laws such as HIPAA. Ensuring data privacy during fine-tuning or testing is of paramount importance. We have not taken this aspect into consideration.

\section{Conclusion}

We have presented our solution submitted to the MEDIQA-Chat shared task for generating clinical notes from doctor-patient dialogues. We evaluated both fine-tuned approaches using models such as CONFIT, GPT-3, RoBERTa, and SciBERT, as well as an approach utilizing GPT-4.

\bibliography{custom}
\bibliographystyle{acl_natbib}

\end{document}